# Robotic Inspection and Characterization of Subsurface Defects on Concrete Structures Using Impact Sounding

EJUP HOXHA, JINGLUN FENG, DIAR SANAKOV,
ARDIAN GJINOFCI and JIZHONG XIAO

**ABSTRACT**

   Impact-sounding (IS) and impact-echo (IE) are well-developed non-destructive evaluation (NDE) methods that are widely used for inspections of concrete structures to ensure the safety and sustainability. However, it is a tedious work to collect IS and IE data along grid lines covering a large target area for characterization of subsurface defects. On the other hand, data processing is very complicated that requires domain experts to interpret the results. To address the above problems, we present a novel robotic inspection system named as Impact-Rover to automate the data collection process and introduce data analytics software to visualize the inspection result allowing regular non-professional people to understand. The system consists of three modules: 1) a robotic platform with vertical mobility to collect IS and IE data in hard-to-reach locations, 2) vision-based positioning module that fuses the RGB-D camera, IMU and wheel encoder to estimate the 6-DOF pose of the robot, 3) a data analytics software module for processing the IS data to generate defect maps. The Impact-Rover hosts both IE and IS devices on a sliding mechanism and can perform move-stop-sample operations to collect multiple IS and IE data at adjustable spacing. The robot takes samples much faster than the manual data collection method because it automatically takes the multiple measurements along a straight line and records the locations. This paper focuses on reporting experimental results on IS. We calculate features and use unsupervised learning methods for analyzing the data. By combining the pose generated by our vision-based localization module and the position of the head of the sliding mechanism we can generate maps of possible defects. The results on concrete slabs demonstrate that our impact-sounding system can effectively reveal shallow defects.

## INTRODUCTION

   The civil infrastructure in the United States of America is reaching pass its life expectancy and the cost of repairs is going to tremendously increase over the next years. ASCE 2021 report card scores America's infrastructure with an average C- grade.

---
Ejup Hoxha, Jinglun Feng, Dr. Jizhong Xiao (Corresponding Author), CCNY Robotics Lab, The City College of New York, 160 Convent Ave, NY 10031, U.S.A.
Diar Sanakov, NYU Tandon School of Engineering, 6 MetroTech, Brooklyn, NY 11201, U.S.A.
Ardian Gjinofci, University of Prishtina "Hasan Prishtina", Rr. "Agim Ramadani", Ndërtesa e Fakulteteve Teknike, p.n., Prishtinë 10000, Republic of Kosovo.

According to this report there are more than 617,000 bridges and 42% of them are over 50 years old, with 7.5% of them considered to be in a poor condition or structurally deficient. Condition of bridges is graded with C, a decrease from C+ from the ASCE 2017 report card. It is critically important to increase the inspection frequency so we can detect and repair the defects before any additional deterioration damages these structures beyond the repair point. The inspection of infrastructure is a time-consuming, expensive and potentially a dangerous process. Inspectors use NDT methods like impact-echo [1], acoustic inspection [2], ultrasonic [3], GPR [4], visual inspection [5] etc., to detect defects like cracks, delamination, voids, rebar corrosion etc. GPR [15][16], is one of the most widely used methods in detecting defects like voids, delamination, buried objects and metallic rebars [6], corrosion [7] etc. Impact-echo method is used to detect delamination, cracks and voids in concrete structures [1]. However, impact-echo although successful still lacks any good result for shallow delamination [8].

The focus of this paper is detection of shallow defects using automated robotic impact-sounding and generation of high-resolution maps. First, an automated robotic solution with wall-climbing capabilities is developed. This automated solution drastically decreases the necessary time to collect data, improves the data consistency so all data points are collected under very similar condition, and finally frees the inspectors from the tedious task of collecting data manually. Second, we use FFT [9] to extract features out of the raw sound files. We show that by using other features other than just frequency peaks, we can use very noisy sound recordings recorded with an affordable microphone, without any physical filter required, and are able to detect voids and delamination in the concrete slab. This paper is potentially impactful, because we use K-Means [10] and Spectral Clustering [26] to analyze the data and we get very good results.

## ROBOTIC IMPACT-SOUNDING SYSTEM

### Robotic System and Software

We developed a robotic inspection system with vertical mobility which is called Impact-Rover (see Figure 1). The robot consists of two DC brushless motors for differential drive locomotion, suction motor for climbing on concrete walls, a solenoid impactor on a sliding mechanism that is controlled by a stepper motor, specially designed embedded system control board, accelerometer with data acquisition module for IE, a microphone for IS, RealSense D435i RGBD camera for localization and Intel NUC on-board computer. Impact-Rover can collect data automatically in a pre-defined area.

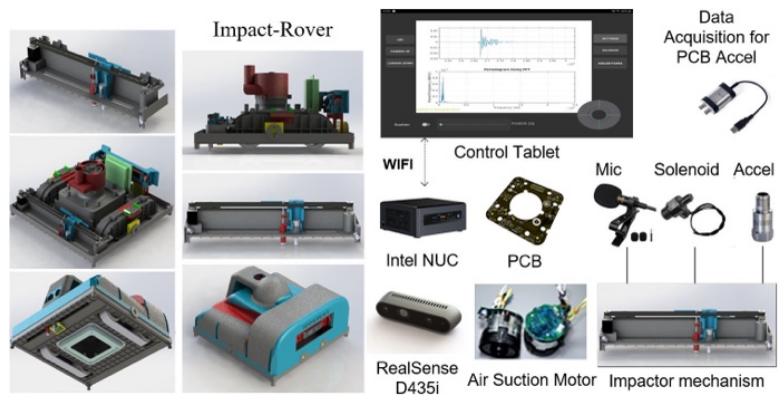

Figure 1. Impact-Rover robot and hardware system.

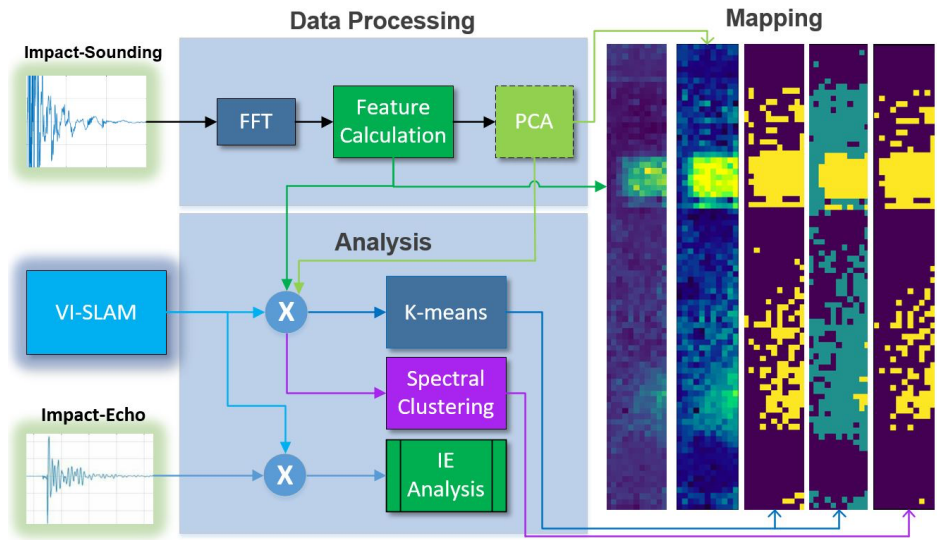

Figure 2. Software system diagram. IS data is transformed to Fourier Domain, then features are calculated. We generate feature maps and perform PCA. Both PCA results and features go into clustering algorithms which generate segmented maps. PCA maps are also very informative to define the cluster properties.

**Impact-Sounding and Impact-Echo Modules**

The impact-sounding module consist of an Olson Solenoid for generating the impact. This solenoid can tap the surface and provide 0.011 to 0.019 Joules per hit. Sound data are recorded using 3.5 mm standard microphone. IS and IE devices both rest on the sliding mechanism to collect data at an adjustable spacing.

The IE module consist of three important parts. An accelerometer from PCB Inc. with sensitivity of 100 mV/g, frequency range ($\pm 3$ dB) 0.5 to 20 kHz. Digital ICP Signal Conditioner Model 485B39 which connects through USB port and is used to transfer the data to the on-board computer. The servo motor is used to lower the IE transducer up and down to be in contact with the concrete surface for taking measurements.

**Localization Module**

In order to automate the data collection process, we need to determine the location of the robot with respect to the starting point and displacement of IE and IS sensors on the sliding mechanism. This way we can tag each measurement with position information. We use RealSense RGBD camera D435i to estimate the pose [11][12][13][14] and then fuse it with wheel's encoders to improve the localization accuracy of the robot.

877

# IMPACT-SOUNDING BASED INSPECTION

## Data Processing and Feature Engineering

Due to the nature of the data and from our previous research experience [12][17][18], we decided to use discrete Fourier Transform [19] or frequency domain analysis. This method decomposes the signal $x(n)$, where $n = 1, 2, ..., N - 1$ into spectral components and provides frequency information (see Equation 1). Thus, inputting a sound signal into Fourier Transform process, as output we get a function in the complex domain. This function represents the amount of each frequency present in the original signal. We perform Fast Fourier Transform (FFT) [20] on all impact-sounding data we collected. For each sound signal we obtain the pair $\{a_k, f_k\}$, where $f_k$ is the frequency vector and $a_k$ corresponds to the amplitude vector, $k = 1, 2, ..., M$.

Feature extracting is the process of extracting non-redundant, abstract or meaningful information out of the data which better represents an underlying problem to the predictive models. Difference between sound signals collected over defective and non-defective concrete are not obvious in time-domain (see Figure 3 a, b and c). On the other hand, FFT signal shows a distinguishable pattern in energy (see Figure 3 d, e and f) i.e., magnitude of energy is lower over defects because they absorb impact energy. Our feature extraction method is inspired by [21], where we use FFT while they use Wavelet Transform [22]. Feature calculations is defined with the following expressions (see Equations 2 to 7 and Table I). These types of features capture different properties of the sound data. From mathematics we know that moments are quantitative measures related to the shape of the function. For these reasons we use spectral moments as features.

$$S(f_k) = \sum_{n=0}^{N-1} x(n) \cdot e^{-\frac{j2\pi}{N} f_k n} \tag{1}$$

$$E = \sum_{i=1}^{N} x_i^2 \tag{2}$$

$$P = \sum_{i=1}^{M} a_i \tag{3}$$

$$M_1 = \sum_{i=1}^{M} \frac{a_i \cdot f_i}{P} \tag{4}$$

TABLE I. FEATURE INFORMATION.

| ID | Equation | Feature Name |
|----|----------|--------------|
| 1 | 2 | Energy |
| 2 | 3 | Power |
| 3 | 4 | First Spectral Moment (Mean) |
| 4 | 5 | Second Spectral Moment (Variance) |
| 5 | 6 | Third Spectral Moment (Skewness) |
| 6 | 7 | Fourth Spectral Moment (Kurtosis) |

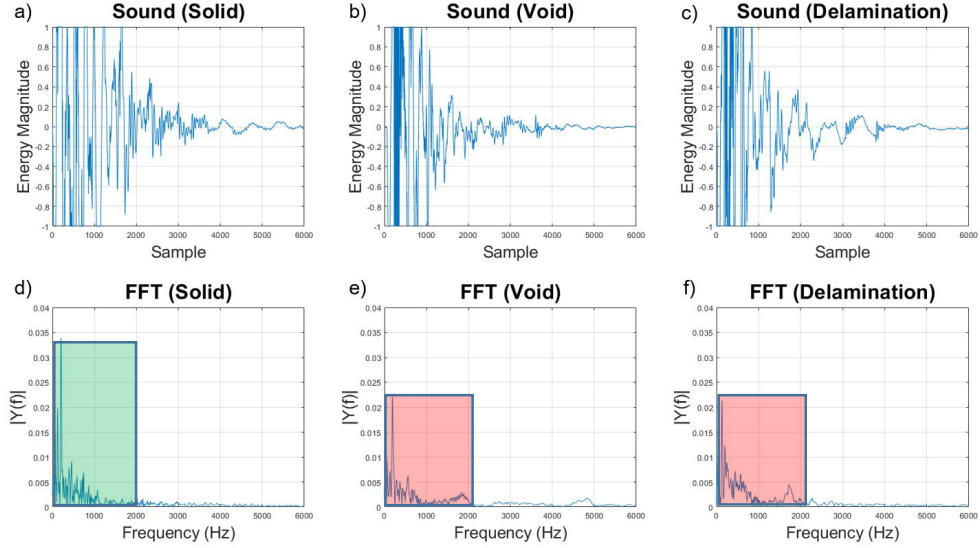

Figure 3. Sample of collected sound data over: a) solid concrete, b) void and c) delamination. Fourier Transform of respective sound signals: d) solid, e) void and f) delamination. Note the difference in energy levels, visible in frequency domain.

$$M_2 = \sum_{i=1}^{M} \frac{a_i \cdot (f_i - M_1)^2}{P} \tag{5}$$

$$M_3 = \sum_{i=1}^{M} \frac{a_i \cdot (f_i - M_1)^3}{P \cdot M_2^3} \tag{6}$$

$$M_3 = \sum_{i=1}^{M} \frac{a_i \cdot (f_i - M_1)^4}{P \cdot M_2^4} \tag{7}$$

$$x_i = f(E_i, P_i, M_{1i}, M_{2i}, M_{3i}, M_{4i}) \tag{8}$$

$$X = (x_1, x_2, \ldots, x_N)^T \tag{9}$$

## Unsupervised clustering and PCA

In this work we use unsupervised classification to classify non-defective and defective concrete. We use K-means [10], which is one of the most famous unsupervised clustering methods. Given the matrix of features $X$ (see Equation 9), this algorithm partitions all observations $(x_1, x_2, \ldots, x_N)$ into $k \leq N$ sets $S^i = (S_1^i, S_2^i, \ldots, S_k^i)$. Given an initial set of centroids $m_{1:k}^1$ randomly initialized, algorithm [24] alternates between two steps. First, calculates the least squared Euclidean distance for each observation and assigns it to the closest centroid (see Equation 10). Second, algorithm recalculates the centroids for updated clusters (see Equation 11). Algorithm ends when $m_{1:k}^t \approx m_{1:k}^{t+1}$. We use Silhouette Coefficient [23] to determine the number of clusters.

$$S_i^t = \{x_n : \|x_p - m_i^t\|^2 \leq \|x_p - m_j^t\|^2 \ \forall j, 1 \leq j \leq k\} \tag{10}$$

$$m_i^{t+1} = \frac{1}{|S_i^t|} \sum_{x_j \in S_i^t} x_j \tag{11}$$

Additionally, we use Principal Component Analysis (PCA) [25] for defect analysis and dimension reduction of feature matrix $X$. PCA takes as input the matrix of data and calculates eigenvalues and eigenvectors, it only uses significant eigenvectors to recreate the data. We use this method in two ways. First, we use the filtering properties of this method for data processing that goes as input into K-means clustering. Second, we visualize some of the principal components and use image enhancement techniques [27] to improve the results.

## EXPERIMENTS

We collected data in our concrete slab over the area that contains two types of defects, void and delamination. We collected 902 data points (.wav) with a 2cm spacing in x and y directions, over an area 20cm wide and 162cm long. Sounds are recorded using a sample rate of 44.1 kHz and a 16-bit AD. Each data point has an ID, which is tagged with the location information with respect to the starting point.

### Feature Visualization

Features contain meaningful information; it helps to visualize the defects. From visualization (see Figure 5) we can clearly see the void box, while delamination circle is visible, but much less. We use enhancement on feature matrix to suppress the noise (Equation 12), leading to much clearer visualization results. For example, first and second moments are enhanced by squaring them, and power feature is enhanced by increasing it to the third power.

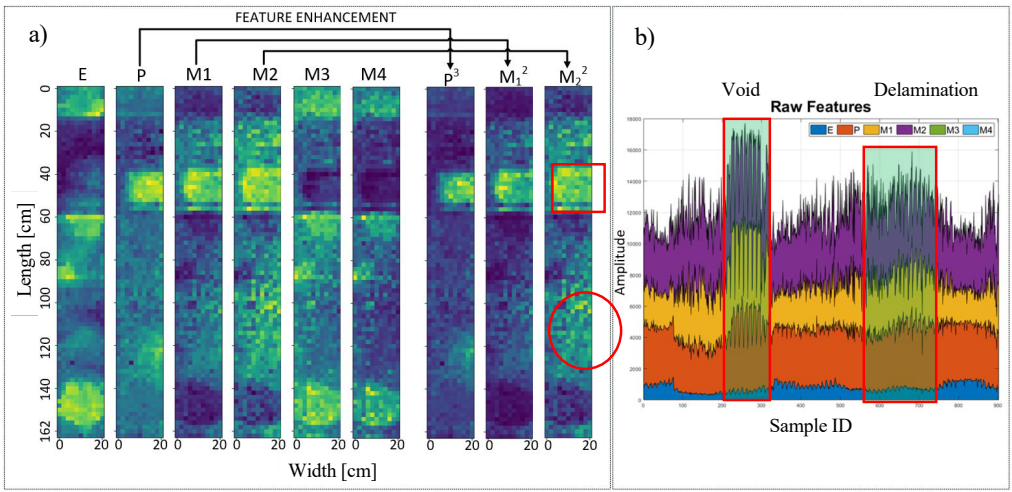

Figure 5. a) Feature maps and feature enhancement. b) Raw features are shown on a 2D plot where the y-axis is the amplitude of raw feature and x-axis represent the sample Id. Void and delamination are visible in this plot.

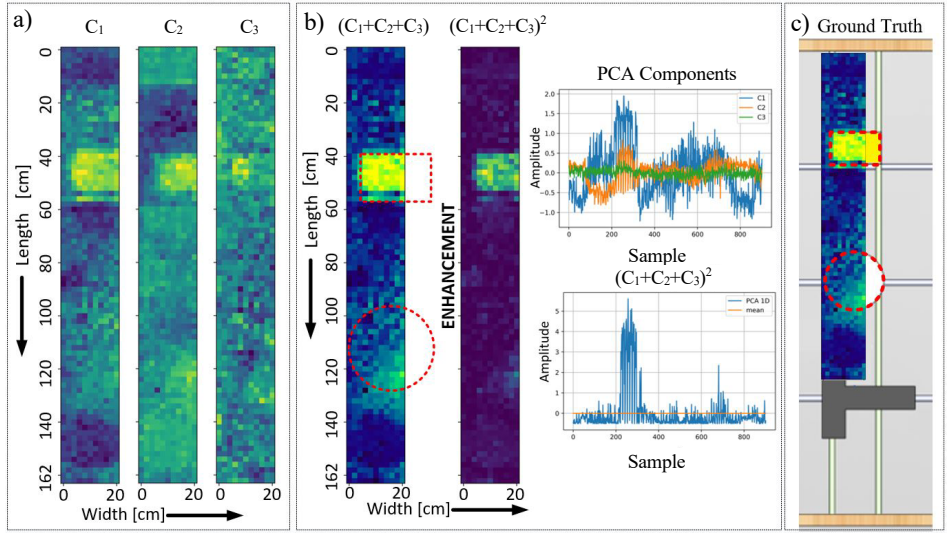

Figure 6. a) PCA components, b) combination maps and c) ground truth. PCA plots and enhanced components plots.

$$\boldsymbol{x}_{enh_i} = [E_i, P_i^3, M_{1i}^2, M_{2i}^2, M_{3i}, M_{4i}] \qquad (12)$$

**PCA Maps**

We use PCA decomposition for the feature matrix $X$ (see Equation 9). For one component (C1) the amount of preserved information is 76.8%, for two components (C1+C2) 96.7% and for three (C1+C2+C3) 98.8%. For visualization we decided to use three components. We visualize all individual components and their linear combination (see Figure 6). They act as three combined predictors with equal weights. From results we can see that using PCA we are able to generate meaningful maps. Void is clearly visible, especially in the enhanced maps. Delamination is weakly detected.

**Unsupervised Clustering Maps**

Up to this point, we generated few types of data that can go into K-means and Spectral Clustering. First type is the feature matrix $\boldsymbol{X}$ (see Equation 8). Second type is enhanced feature matrix, where feature vectors are defined as $\boldsymbol{x}_{enh_i}$ (see Equation 12). Third type is PCA. We use Silhouette coefficient to determine how good the clusters are separated, closer to 1 is better, -1 is worse. Table II shows the values for all cases when using two (defect/no defect) three clusters (void/delamination/no defect). According to the Table II, best clustering is achieved by Spectral Clustering using two clusters.

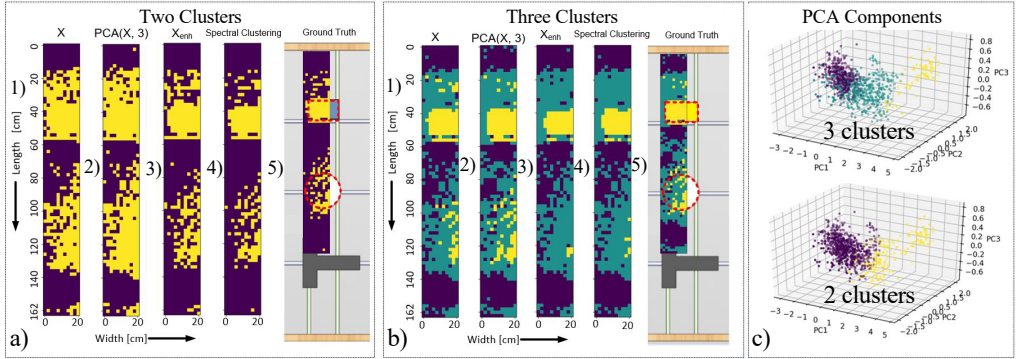

Figure 7. Clustering Results. a) Two clusters result matches with the ground truth and clearly finds the defects. b) Three clusters result shows that concrete around artificial defects is deteriorated. c) Scatter plots of PCA components with cluster information.

When we use $X$ and $PCA(X, 3)$ as input for the K-means, the output is noisy and not very clear (see Figure 7a1, 7a2, 7b1 and 7b2). Feature enhancement improves the results (see Figure 7a1 and 7a3). Spectral Clustering uses the unmodified features $X$ and outputs very similar results with feature enhancement (see Figure 7a3 and 7a4). On the other hand, having three clusters differentiates the void and delamination (see Figure 7b). This figure shows parts of concrete close by artificial defects as weakened or deteriorated. Finally, 3D scatter plots are generated by using PCA decomposition with cluster information (see Figure 7c).

## CONCLUSION

This paper introduces a robotic solution for data collection and an unsupervised learning method for impact-sounding data analysis. The robotic platform is equipped with visual, inertial and encoder fusion for high accuracy localization. It makes data collection much faster and improves data consistency. Data analysis algorithm starts with transforming the sound data into the spectral domain, then we calculate features. We use PCA to generate maps, then use K-means and Spectral Clustering for classification. Results from PCA maps and clustered maps can be used together to reveal defects. Experimental results show the effectiveness of our proposed method.

## ACKNOLEDGEMENT

Financial support for this study was provided by the U.S. Department of Transportation under Grant No. 69A3551747126 through INSPIRE UTC. J. Xiao has significant financial interest in InnovBot LLC, a company involved in R&D and commercialization of the technology.

TABLE II. SILHOUETTE COEFFICIENT FOR CLUSTERING.

| Clusters | $X$ | $PCA(X, 3)$ | $X_{enh}$ | Spectral Cls. |
|---|---|---|---|---|
| 2 | 0.4263 | 0.435 | 0.4622 | **0.4785** |
| 3 | 0.4105 | 0.413 | 0.4732 | 0.4435 |

# REFERENCES

1. Mary J. Sansalone, William B. Streett. 1997. *Impact-Echo: Nondestructive Evaluation of Concrete and Masonry*. 1-339.
2. A. Watanabe, J. Even, L. Y. Morales, and C. Ishi. 2015. "Robot-assisted acoustic inspection of infrastructures-cooperative hammer sounding inspection", in *IEEE/RSJ International Conference on Intelligent Robots and Systems (IROS)*. IEEE, 2015, pp. 5942–5947.
3. Josef Krautkrämer, Herbert Krautkrämer. 1990. *Ultrasonic testing of materials*, 4th fully rev. ed. Berlin; New York: Springer-Verlag. ISBN 3-540-51231-4.
4. A. Annan. 2005. "Ground-penetrating radar", in *Near-surface geophysics, Society of Exploration of Geophysics*, pp. 357-438.
5. L. Yang, B. Li, W. Li, H. Brand, B. Jiang and J. Xiao. 2020. "Concrete defects inspection and 3D mapping using CityFlyer quadrotor robot," in *IEEE/CAA Journal of Automatica Sinica*, vol. 7, no. 4, pp. 991-1002, July 2020, doi: 10.1109/JAS.2020.1003234.
6. J. Feng, L. Yang, E. Hoxha, D. Sanakov, S. Sotnikov and J. Xiao. 2021. "GPR-based Model Reconstruction System for Underground Utilities Using GPRNet," in *IEEE International Conference on Robotics and Automation (ICRA)*, 2021, pp. 845-851, doi: 10.1109/ICRA48506.2021.9561355.
7. Tešić, Ksenija et al. 2021. "Non-Destructive Corrosion Inspection of Reinforced Concrete Using Ground-Penetrating Radar: A Review." *Materials (Basel, Switzerland)* vol. 14,4 975. 19 Feb. 2021, doi:10.3390/ma14040975.
8. Jinying Zhu and John S. Popovics. 2007. "Imaging Concrete Structures Using Air-Coupled Impact-Echo", *Journal of Engineering Mechanics*, Vol. 133, pp. 628-640, 2007, doi: 10.1061/(ASCE) 0733-9399.
9. E. O. Brigham and R. E. Morrow. 1967. "The fast Fourier transform," in *IEEE Spectrum*, vol. 4, no. 12, pp. 63-70, Dec. 1967, doi: 10.1109/MSPEC.1967.5217220.
10. Lloyd, Stuart P. 1982. "Least squares quantization in PCM." *Information Theory, IEEE Transactions* on 28.2 (1982): 129-137.
11. I. Dryanovski, R. G. Valenti, and J. Xiao. 2013. "Fast visual odometry and mapping from rgb-d data", in 2013 *IEEE International Conference on Robotics and Automation*. IEEE, 2013, pp. 2305–2310.
12. Jinglun Feng, Hua Xiao, Ejup Hoxha, Haiyan Wang, Yifeng Song, Liang Yang, Jizhong Xiao. 2021. "Automatic Impact-sounding Acoustic Inspection of Concrete Structure", *10th International Conference on Structural Health Monitoring of Intelligent Infrastructure Advanced Research and Real-world Applications (SHMII-10)*, 30 June – 2 July 2021, Porto, Portugal.
13. Solà, J., 2017. "Quaternion kinematics for the error-state Kalman filter", arXiv e-prints, 2017.
14. Kalman, R. 1960. "A New Approach to Linear Filtering and Prediction Problems". *ASME Journal of Basic Engineering*, 82, 35-45. http://dx.doi.org/10.1115/1.3662552.
15. Jinglun Feng, Liang Yang, Haiyan Wang, Yingli Tian and Jizhong Xiao. 2021. "Subsurface Pipes Detection Using DNN-based Back Projection on GPR Data", *Proceedings of the IEEE/CVF Winter Conference on Applications of Computer Vision (WCACV2021)*, Jan 5~9, 2021, pp. 266-275.
16. Jinglun Feng, Liang Yang, Haiyan Wang, Yifeng Song, Jizhong Xiao. 2020. "GPR-based Subsurface Object Detection and Reconstruction Using Random Motion and DepthNet", *IEEE International Conference on Robotics and Automation (ICRA2020)*, pp. 7035~7041, May 31 – August 31, 2020.
17. B. Li, K. Ushiroda, L. Yang, Q. Song, and J. Xiao. 2017. "Wall-climbing robot for non-destructive evaluation using impact-echo and metric learning svm," *International Journal of Intelligent Robotics and Applications*, vol. 1, no. 3, pp. 255–270, 2017.
18. Li, J. Cao, J. Xiao, X. Zhang, and H. Wang. 2014. "Robotic impact-echo non-destructive evaluation based on fft and svm," in *Proceeding of the 11th World Congress on Intelligent Control and Automation*. IEEE, 2014, pp. 2854–2859.
19. W. Jenkins and M. Desai. 1986. "The discrete frequency Fourier transform" in *IEEE Transactions on Circuits and Systems*, vol. 33, no. 7, pp. 732-734, July 1986, doi: 10.1109/TCS.1986.1085978.
20. E. O. Brigham and R. E. Morrow. 1967. "The fast Fourier transform," in *IEEE Spectrum*, vol. 4, no. 12, pp. 63-70, Dec. 1967, doi: 10.1109/MSPEC.1967.5217220.

21. Zhang, Jing-Kui et al. 2016. "Concrete Condition Assessment Using Impact-Echo Method and Extreme Learning Machines." *Sensors (Basel, Switzerland)* vol. 16,4 447. 26 Mar. 2016, doi:10.3390/s16040447.
22. Daubechies I. 1992. *"Ten Lectures on Wavelets"*. Society for Industrial and Applied Mathematics; Philadelphia, PA, USA: 1992.
23. Peter J. Rousseeuw, "Silhouettes: A graphical aid to the interpretation and validation of cluster analysis", *Journal of Computational and Applied Mathematics*, Volume 20, 1987, Pages 53-65, ISSN 0377-0427, https://doi.org/10.1016/0377-0427(87)90125-7.
24. Protter, Murray H.; Morrey, Jr., Charles B. 1970. *College Calculus with Analytic Geometry (2nd ed.)*, Reading: Addison-Wesley, LCCN 76087042, p. 521.
25. Karl Pearson F.R.S. 1901. LIII. *On lines and planes of closest fit to systems of points in space*. The London, Edinburgh, and Dublin Philosophical Magazine and Journal of Science, 2(11), 559–572. https://doi.org/10.1080/14786440109462720.
26. Ulrike von Luxburg. 2007. "A tutorial on spectral clustering" in *Statistical Computing* 17, 395–416 (2007). https://doi.org/10.1007/s11222-007-9033-z.
27. Warf, B. 2010. "Image enhancement". In *Encyclopedia of geography* (Vol. 1, pp. 1533-1533). SAGE Publications, Inc., https://www.doi.org/10.4135/9781412939591.n610.